%% file: main.tex
\documentclass[11pt]{article}
\usepackage{acl2014}
\usepackage{times}
\usepackage{url}
\usepackage{latexsym}
\usepackage{amsmath}
\usepackage{multirow}
\usepackage[normalem]{ulem}
\usepackage{color}
\usepackage[usenames,dvipsnames]{xcolor}
\usepackage{enumitem}
\usepackage[lined,boxed,commentsnumbered]{algorithm2e}
\usepackage{graphicx}
\usepackage{epstopdf}
\usepackage{caption}
\usepackage{subcaption}
\usepackage{mdwlist}
\usepackage{enumitem}

\title{A Piece of My Mind: A Sentiment Analysis Approach\\ for Online Dispute Detection}
\author{Lu Wang \\
  Department of Computer Science \\
  Cornell University \\
  Ithaca, NY 14853 \\
  {\tt luwang@cs.cornell.edu} \\\And
  Claire Cardie \\
  Department of Computer Science \\
  Cornell University \\
  Ithaca, NY 14853 \\
  {\tt cardie@cs.cornell.edu} \\}

\begin{document}
\maketitle

\input{abstract.tex}

\section{Introduction}
\label{sec:intro}
\input{intro.tex}

\paragraph{Additional Related Work.}
\label{sec:related}
\input{related.tex}
\section{Data Construction: A Dispute Corpus}
\label{sec:data}
\input{data.tex}

\section{Sentence-level Sentiment Prediction}
\label{sec:model}
\input{model.tex}

\section{Online Dispute Detection}
\label{sec:dispute}
\input{dispute.tex}

\vspace{-1mm}
\section{Conclusion}
\vspace{-1mm}
\input{conclusion.tex}

\noindent{}
{\bf Acknowledgments} {\fontsize{9}{8}\selectfont We heartily thank the Cornell NLP Group, the reviewers, and Yiye Ruan for helpful comments. We also thank Emily Bender and Mari Ostendorf for providing the AAWD dataset. This work was supported in part by NSF grants IIS-0968450 and IIS-1314778, and DARPA DEFT Grant FA8750-13-2-0015. The views and conclusions contained herein are those of the authors and should not be interpreted as necessarily representing the official policies or endorsements, either expressed or implied, of NSF, DARPA or the U.S. Government.}

\bibliographystyle{acl}

\end{document}

%% file: abstract.tex
\begin{abstract}
\fontsize{10}{11}\selectfont
We investigate the novel task of \textit{online dispute detection} and propose a sentiment analysis solution to the problem: we aim to identify the sequence of sentence-level sentiments expressed during a discussion and to use them as features in a classifier that predicts the DISPUTE/NON-DISPUTE label for the discussion as a whole.
We evaluate dispute detection approaches on a newly created corpus of Wikipedia Talk page disputes and find that classifiers that rely on our sentiment tagging features outperform those that do not. The best model achieves a very promising F1 score of 0.78 and an accuracy of 0.80.
\end{abstract}

%% file: intro.tex
As the web has grown in popularity and scope, so has the promise of collaborative information environments for the joint creation and exchange of knowledge~\cite{jones:2000,sack:2005}. Wikipedia, a wiki-based online encyclopedia, is arguably the best example: its distributed editing environment allows readers to collaborate as content editors and has facilitated the production of over four billion articles\footnote{\scriptsize \url{http://en.wikipedia.org}} of surprisingly high quality~\cite{giles:2005} in English alone since its debut in 2001.  
%

Existing studies of collaborative knowledge systems have shown, however, that the quality of the generated content (e.g.\ an encyclopedia article) is highly correlated with the effectiveness of the online collaboration \cite{kittur:2008:HWC,kraut-resnick:2012}; fruitful collaboration, in turn, inevitably requires dealing with the disputes and conflicts that arise \cite{Kittur2007HSS}.
%
%
Unfortunately, human monitoring of the often massive social media and collaboration sites to detect, much less mediate, disputes is not feasible.

In this work, we investigate the heretofore novel task of \textit{dispute detection in online discussions}. 
Previous work in this general area has analyzed dispute-laden content to discover features correlated with conflicts and disputes~\cite{Kittur2007HSS}. Research focused primarily on cues derived from the edit history of the jointly created content (e.g.\ the number of revisions, their temporal density \cite{Kittur2007HSS,abs-1202-3643}) and relied on  small numbers of manually selected discussions known to involve disputes.  
In contrast, we investigate methods for the automatic detection, i.e.\ prediction, of discussions involving disputes.  We are also interested in understanding whether, and which, linguistic features of the discussion are important for dispute detection.

Drawing inspiration from studies of human mediation of online conflicts (e.g.\ \newcite{BillingsW10}, \newcite{Kittur2007HSS}, \newcite{kraut-resnick:2012}), we hypothesize that effective methods for dispute detection should take into account the sentiment and opinions expressed by participants in the collaborative endeavor. 
As a result, we propose a sentiment analysis approach for online dispute detection that identifies the sequence of sentence-level sentiments (i.e.\ very negative, negative, neutral, positive, very positive) expressed during the discussion and uses them as features in a classifier that predicts the {\sc dispute}/{\sc non-dispute} label for the discussion as a whole. 
Consider, for example, the snippet in Figure~\ref{fig:discussion_example} from the Wikipedia Talk page for the article on Philadelphia; it discusses the choice of a picture for the article's ``infobox".  The sequence of almost exclusively negative statements provides evidence of a dispute in this portion of the discussion.

\begin{figure}[th]
\hspace{-3mm}
    {\fontsize{7.5}{8}\selectfont
    \begin{tabular}{|p{78mm}|}
    \hline
    1-\textbf{Emy111}: I think everyone is forgetting that my previous image was the lead image for well over a year! ...\\ 
    $>$ \textbf{Massimo}: I'm sorry to say so, but it is grossly over processed...\\
    2-\textbf{Emy111}: 
    i'm glad you paid more money for a camera than I did. {\color{blue}congrats...} i appreciate your constructive criticism. {\color{blue}thank you.}\\
    $>$ \textbf{Massimo}: 
    I just want to have the best picture as a lead for the article ...\\
    3-\textbf{Emy111}: Wow, I am really enjoying this photography debate... 
    {\color{blue}[so don't make assumptions you know nothing about.]$_{\textbf{NN}}$ [Really, grow up.]$_{\textbf{N}}$ [If you all want to complain about Photoshop editing, lets all go buy medium format film cameras, shoot film, and scan it, so no manipulation is possible.]$_{\textbf{O}}$ [Sound good?]$_{\textbf{NN}}$}\\
    $>$ \textbf{Massimo}: ... I do feel it is a pity, that you turned out to be a sore loser... 
    \\ \hline
	\end{tabular}
	}
	\vspace{-4mm}
    \caption{\fontsize{8}{9}\selectfont From the Wikipedia Talk page for the article ``Philadelphia". Omitted sentences are indicated by ellipsis. Names of editors are in \textbf{bold}. The start of each set of related turns is numbered; ``$>$" is an indicator for the reply structure.}
    \label{fig:discussion_example}
\end{figure}

Unfortunately, sentence-level sentiment tagging for this domain is challenging in its own right due to the less formal, often ungrammatical, language and the dynamic nature of online conversations. 
``\textit{Really, grow up}" (segment 3) should presumably be tagged as a negative sentence as should the sarcastic sentences ``\textit{Sounds good?}" (in the same turn) and ``\textit{congrats}" and ``\textit{thank you}" (in segment 2). We expect that these, and other, examples will be difficult for the sentence-level classifier unless the discourse context of each sentence is considered.
Previous research on sentiment prediction for online discussions, however, focuses on turn-level predictions~\cite{hahn-ladner-ostendorf:2006:HLT-NAACL06-Short,Yin:2012:ULG:2392963.2392978}.\footnote{A notable exception is~\newcite{Hassan:2010:WAI:1870658.1870779}, which identifies sentences containing ``attitudes" (e.g.\ opinions), but does not distinguish them w.r.t.\ sentiment. Context information is also not considered.}
As the first work that predicts sentence-level sentiment for online discussions, we investigate isotonic Conditional Random Fields (CRFs)~\cite{Mao+Lebanon:07a} for the sentiment-tagging task as they preserve the advantages of the popular CRF-based sequential tagging models~\cite{Lafferty:2001:CRF} while providing an efficient mechanism for encoding domain knowledge --- in our case, a sentiment lexicon --- through isotonic constraints on model parameters.  
%



We evaluate our dispute detection approach using a newly created corpus of discussions from Wikipedia Talk pages (3609 disputes, 3609 non-disputes).\footnote{The talk page associated with each article records conversations among editors about the article content and allows editors to discuss the writing process, e.g.\ planning and organizing the content.} 
We find that classifiers that employ the learned sentiment features outperform others that do not. The best model achieves a very promising F1 score of 0.78 and an accuracy of 0.80 on the Wikipedia dispute corpus. To the best of our knowledge, this represents the first computational approach to automatically identify online disputes on a dataset of scale.


%% file: related.tex
%
%
Sentiment analysis has been utilized as a key enabling technique in a number of conversation-based applications. Previous work mainly studies the attitudes in spoken meetings~\cite{GalleyEtal,hahn-ladner-ostendorf:2006:HLT-NAACL06-Short} or broadcast conversations~\cite{Wang:2011:DAD} using variants of Conditional Random Fields~\cite{Lafferty:2001:CRF}  
and predicts sentiment at the turn-level, while our predictions are made for each sentence.

%% file: data.tex
We construct the first dispute detection corpus to date; it consists of dispute and non-dispute discussions from Wikipedia Talk pages. 

\noindent
\textbf{Step 1: Get Talk Pages of Disputed Articles.}
Wikipedia articles are edited by different editors. If an article is observed to have disputes on its \textit{talk page}, editors can assign dispute tags to the article to flag it for attention. 
In this research, we are interested in talk pages whose corresponding articles are labeled with the following tags: {\sc disputed}, {\sc totallydisputed}, {\sc disputed-section}, {\sc totallydisputed-section}, {\sc pov}. 
The tags indicate that an article is disputed, or the neutrality of the article is disputed 
({\sc pov}).
 
We use the 2013-03-04 Wikipedia data dump, and extract talk pages for articles that are labeled with dispute tags by checking the revision history. This results in 19,071 talk pages.

\noindent
\textbf{Step 2: Get Discussions with Disputes.}
Dispute tags can also be added to \textit{talk pages} themselves. Therefore, 
in addition to the tags mentioned above, we also consider the ``Request for Comment" 
({\sc rfc}) tag on talk pages. According to Wikipedia\footnote{\scriptsize \url{http://en.wikipedia.org/wiki/Wikipedia:Requests_for_comment}}, 
{\sc rfc} is used to request outside opinions concerning the disputes. 

3609 discussions are collected with dispute tags found in the revision history. We further classify dispute discussions into three subcategories: {\sc Controversy}, {\sc Request for Comment (RFC)}, and {\sc Resolved} based on the tags found in discussions (see Table~\ref{tab:dispute_tag}). 
The numbers of discussions for the three types are 42, 3484, and 105, respectively.
Note that dispute tags only appear in a small number of articles and talk pages. There may exist other discussions with disputes. 

\vspace{-2mm}
\begin{table}[ht]
    {\fontsize{7.5}{8}\selectfont
	\begin{tabular}{|l|l|}
    \hline
	\textbf{Dispute Subcategory}& \textbf{Wikipedia Tags on Talk pages}\\ \hline
	Controversy & {\sc Controversial}, {\sc totallydisputed},  \\
				& {\sc Disputed}, {\sc Calm talk}, {\sc POV} \\ \hline
	Request for Comment & {\sc rfc} \\ \hline
	Resolved & Any tag from above + {\sc Resolved}\\ \hline
	\end{tabular}
	}
	\vspace{-4mm}
    \caption{\fontsize{8}{9}\selectfont  Subcategory for disputes with corresponding tags. Note that each discussion in the {\sc Resolved} class has more than one tag.}
    \vspace{-3mm}
    \label{tab:dispute_tag}
\end{table}

\noindent
\textbf{Step 3: Get Discussions without Disputes.}
Likewise, we collect non-dispute discussions from pages that are never tagged with disputes.  
We consider non-dispute discussions with at least 3 distinct speakers and 10 turns. 3609 discussions are randomly selected with this criterion. The average turn numbers for dispute and non-dispute discussions are $45.03$ and $22.95$, respectively.


%% file: model.tex

This section describes our sentence-level sentiment tagger, from which we construct features for dispute detection (Section~\ref{sec:dispute}).

Consider a discussion comprised of sequential turns; 
each turn consists of a sequence of sentences. Our model takes as input the sentences $\mathbf{x}=\{x_{1}, \cdots, x_{n}\}$ from a single turn, and outputs the corresponding sequence of sentiment labels $\mathbf{y}=\{y_{1}, \cdots, y_{n}\}$, where $y_{i}\in \mathcal{O},  \mathcal{O}=\{\mathrm{NN, N, O, P, PP}\}$. The labels in $\mathcal{O}$ represent very negative (NN), negative (N), neutral (O), positive (P), and very positive (PP), respectively. 




Given that traditional Conditional Random Fields (CRFs)~\cite{Lafferty:2001:CRF} ignore the ordinal relations among sentiment labels, we choose \textit{isotonic CRFs}~\cite{Mao+Lebanon:07a} for sentence-level sentiment analysis as they can enforce monotonicity constraints on the parameters consistent with the ordinal structure and domain knowledge (e.g.\ word-level sentiment conveyed via a lexicon). 
Concretely, we take a lexicon $\mathcal{M}=\mathcal{M}_{p}\cup \mathcal{M}_{n}$, where $\mathcal{M}_{p}$ and $\mathcal{M}_{n}$ are two sets of features (usually words) identified as strongly associated with positive and negative sentiment. Assume $\mu_{ \langle \sigma, w \rangle }$ encodes the weight between label $\sigma$ and feature $w$, for each feature $w \in \mathcal{M}_{p}$; then the isotonic CRF enforces $\sigma \leq \sigma^{\prime} \Rightarrow \mu_{\langle \sigma, w \rangle} \leq \mu_{\langle \sigma^{\prime}, w \rangle}$.  
For example, when we observe ``totally agree" in the training data, the feature parameter for $\mu_{\langle \mathrm{PP}, \mathrm{totally~agree} \rangle }$ is likely to increase. Similar constraints are defined on $\mathcal{M}_{n}$.

Our lexicon is built by combining MPQA~\cite{Wilson:2005:RCP}, General Inquirer~\cite{stone66}, and SentiWordNet~\cite{Esuli2006sentiwordnet} lexicons. Words with contradictory sentiments are removed. We use the features in Table~\ref{tab:feature_isocrf} for sentiment prediction. 

\begin{table}
	\centering
    {\fontsize{7.5}{8}\selectfont
    \setlength{\baselineskip}{0pt}
    \begin{tabular}{|l|l|}
    \hline
    
    \underline{\bf Lexical Features} 	& \underline{{\bf Syntactic/Semantic Features}}\\ 
    - unigram/bigram 					&    	- unigram with POS tag\\
    - number of words all uppercased  		&     	- dependency relation\\
    - number of words 						& \underline{{\bf Conversation Features}}\\ 
    
    \underline{{\bf Discourse Features}} & 	- quote overlap with target\\    
	- initial uni-/bi-/tri-gram 			& 	- TFIDF similarity with target \\
	- repeated punctuations 				& 		(remove quote first)\\
	- hedging phrases collected from		&     \underline{{\bf Sentiment Features}}\\ 
	\newcite{Farkas:2010:CST:1870535.1870536}	&     - connective + sentiment words\\
	- number of negators 				&     - sentiment dependency relation\\
										&	 - sentiment words\\

	\hline
    \end{tabular}    
    }
	\vspace{-4mm}
    \caption{\fontsize{8}{9}\selectfont Features used in sentence-level sentiment prediction. Numerical features are first normalized by standardization, then binned into 5 categories.}
    \vspace{-3mm}
    \label{tab:feature_isocrf}
\end{table}


\noindent
\textbf{Syntactic/Semantic Features.} 
We have two versions of dependency relation features, the original form and a form that generalizes a word to its POS tag, e.g.\ ``nsubj(wrong, you)" is generalized to ``nsubj(\texttt{ADJ}, you)" and ``nsubj(wrong, \texttt{PRP})". 

\noindent
\textbf{Discourse Features.}
We extract the initial unigram, bigram, and trigram of each utterance as discourse features~\cite{Hirschberg:1993:ESD:972487.972490}. 


\noindent
\textbf{Sentiment Features.}
We gather connectives from the Penn Discourse TreeBank~\cite{penntreediscourse} and combine them with any sentiment word that precedes or follows it as new features. Sentiment dependency relations are the dependency relations that include a sentiment word. We replace those words with their polarity equivalents. For example, relation ``nsubj(wrong, you)" becomes ``nsubj(\texttt{SentiWord}$_{neg}$, you)".

%% file: dispute.tex

\subsection{Training A Sentiment Classifier}
\label{subsec:sentiClassifier}

\noindent
\textbf{Dataset.} 
We train the sentiment classifier using the \textit{Authority and Alignment in Wikipedia Discussions (AAWD)} corpus~\cite{Bender:2011:ASA} on a 5-point scale (i.e.\ NN, N, O, P, PP). 
AAWD consists of 221 English Wikipedia discussions with positive and negative alignment annotations. 
%
%
Annotators either label each sentence as positive, negative or neutral, or label the full turn. For instances that have only a turn-level label, we assume all sentences have the same label as the turn. We further transform the labels into the five sentiment labels. Sentences annotated as being a positive alignment by at least two annotators are treated as very positive (PP). If a sentence is only selected as positive by one annotator or obtains the label via turn-level annotation, it is positive (P). Very negative (NN) and negative (N) are collected in the same way. All others are neutral (O). Among all 16,501 sentences in AAWD, 1,930 and 1,102 are labeled as NN and N. 532 and 99 of them are PP and P. The other 12,648 are considered neutral.


\noindent
\textbf{Evaluation.} 
To evaluate the performance of the sentiment tagger, we compare to two baselines. (1) \textbf{Baseline (Polarity)}: a sentence is predicted as positive if it has more positive words than negative words, or negative if more negative words are observed. Otherwise, it is neutral. (2) \textbf{Baseline (Distance)} is extended from~\cite{Hassan:2010:WAI:1870658.1870779}. Each sentiment word is associated with the closest second person pronoun, and a surface distance is computed. An SVM classifier~\cite{Joachims:1999:MLS:299094.299104} is trained using features of the sentiment words and minimum/maximum/average of the distances.

We also compare with two state-of-the-art methods that are used in sentiment prediction for conversations: (1) an SVM (RBF kernel) that is employed for identifying sentiment-bearing sentences~\cite{Hassan:2010:WAI:1870658.1870779}, and (dis)agreement detection~\cite{Yin:2012:ULG:2392963.2392978} in online debates; (2) a 
Linear 
CRF for (dis)agreement identification in broadcast conversations~\cite{Wang:2011:DAD}.

\begin{table}[t]
\centering
    {\fontsize{8}{9}\selectfont
	\begin{tabular}{|l|ccc|}
    \hline
	& \textbf{Pos} & \textbf{Neg} & \textbf{Neutral}\\ \hline
	Baseline (Polarity) & 22.53 & 38.61 & 66.45\\ 
	Baseline (Distance) & 33.75 & 55.79 & 88.97 \\ \hline
	SVM (3-way) & 44.62 & 52.56 & 80.84\\ 
	CRF (3-way) & 56.28 & 56.37& 89.41\\ 
	CRF (5-way) & 58.39& 56.30 & 90.10\\ 
	isotonic CRF & \textbf{68.18} & \textbf{62.53} & 88.87\\ \hline
	\end{tabular}
	}
	\vspace{-3mm}
    \caption{\small F1 scores for positive and negative alignment on Wikipedia Talk pages (AAWD) using 5-fold cross-validation. In each column, \textbf{bold} entries (if any) are statistically significantly higher than all the rest.
    We also compare with an SVM and linear CRF trained with three classes (3-way).
    Our model based on the isotonic CRF produces significantly better results than all the other systems.} 
    \vspace{-3mm}
	\label{tab:aawd_main}
\end{table}

We evaluate the systems using standard F1 on classes of positive, negative, and neutral, where samples predicted as PP and P are positive alignment, and samples tagged as NN and N are negative alignment. 
Table~\ref{tab:aawd_main} describes the main results on the AAWD dataset: our isotonic CRF based system significantly outperforms the alternatives for positive and negative alignment detection (paired-$t$ test, $p<0.05$).

\begin{figure*}[t]
  \hspace{-12mm}
  \begin{minipage}[]{0.72\textwidth}
    \includegraphics[width=110mm,height=28mm]{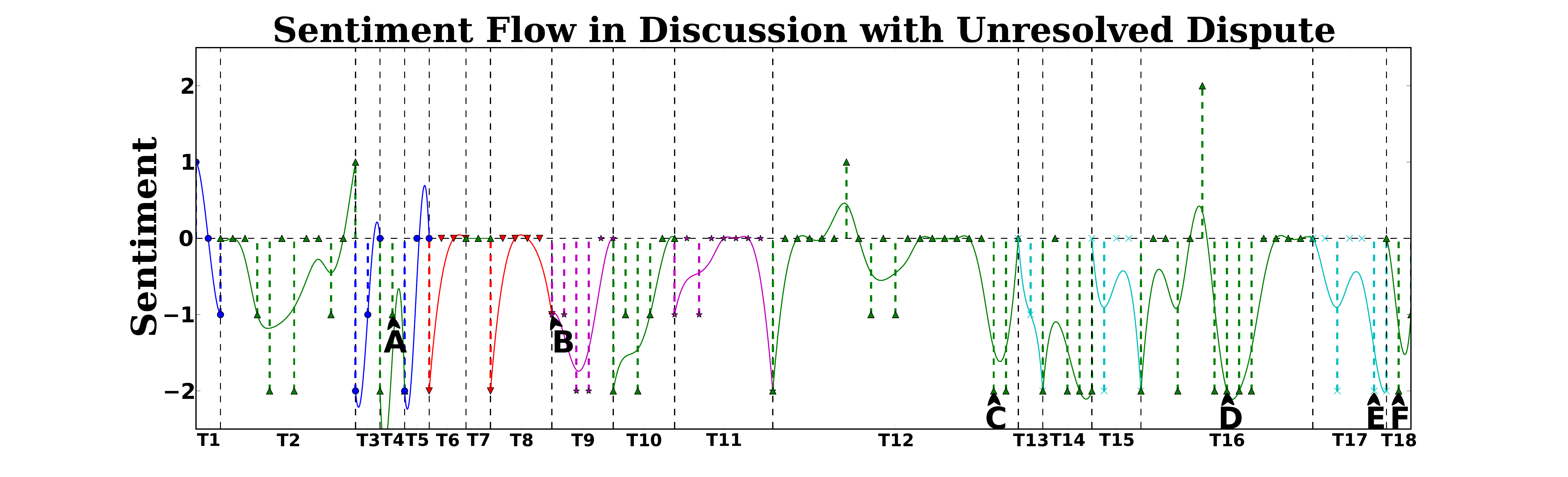}
  \end{minipage}
  \begin{minipage}[]{0.40\textwidth}
    \hspace{-18mm}
    {\fontsize{7}{7}\selectfont
	\begin{tabular}{p{72mm}}
	\underline{\textbf{Sample sentences (sentiment in parentheses)}}\\
	
	\textbf{A}: no, I sincerely plead with you... (N) If not, you are just wasting my time. (NN)\\
	\textbf{B}: I believe Sweet's proposal... is quite silly. (NN)\\
	\textbf{C}: Tell you what. (NN) If you can get two other editors to agree... I will shut up and sit down. (NN)\\
	\textbf{D}: But some idiot forging your signature claimed that doing so would violate. (NN)... Please go have some morning coffee. (O)\\
	\textbf{E}: And I don't like coffee. (NN) Good luck to you. (NN)\\
	\textbf{F}: Was that all? (NN)... I think that you are in error... (N)\\
	
	\end{tabular}
	}
  \end{minipage}
  
  \vspace{-2mm}
  \hspace{-12mm}
  \begin{minipage}[]{0.72\textwidth}
    \includegraphics[width=110mm,height=28mm]{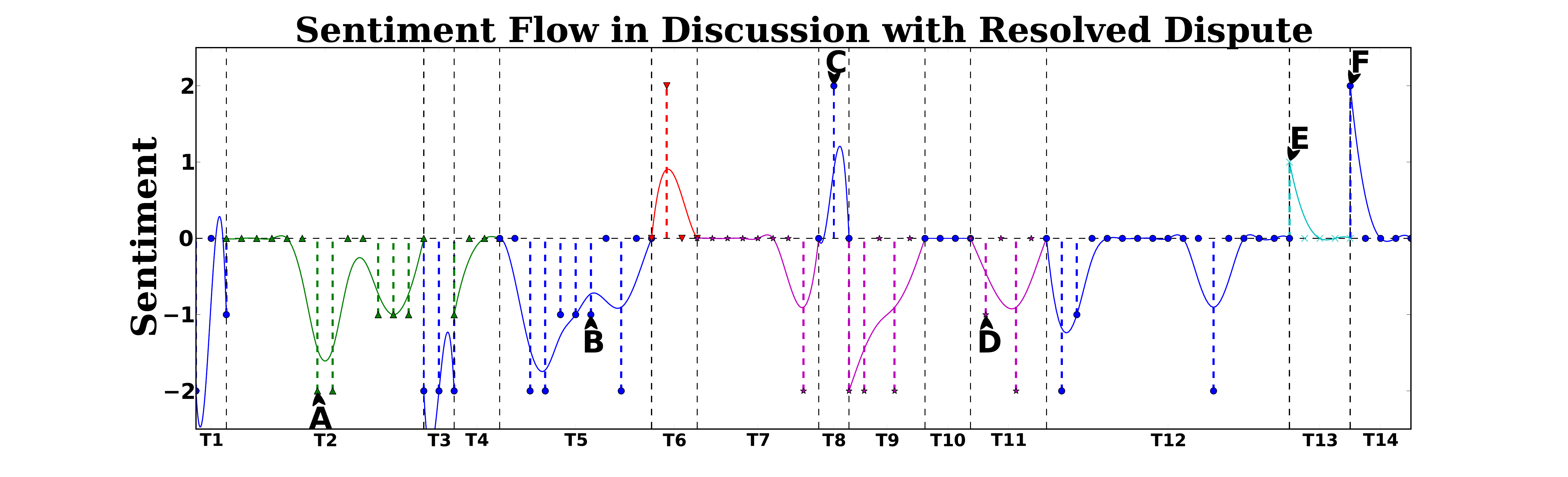}
  \end{minipage}
  \begin{minipage}[]{0.40\textwidth}
    \hspace{-18mm}
    {\fontsize{7}{7}\selectfont
	\begin{tabular}{p{72mm}}
	\textbf{A}: So far so confusing. (NN)...\\
	\textbf{B}: ... I can not see a rationale for the landrace having its own article... (N) With Turkish Van being a miserable stub, there's no such rationale for forking off a new article... (NN)...\\
	\textbf{C}: I've also copied your post immediately above to that article's talk page since it is a great ``nutshell" summary. (PP)\\
	\textbf{D}: Err.. how can the opposite be true... (N)\\
	\textbf{E}: Thanks for this, though I have to say some of the facts floating around this discussion are wrong. (P)\\
	\textbf{F}: Great. (PP) Let's make sure the article is clear on this. (O)\\ \hline
	
	\end{tabular}
	}
  \end{minipage}
  
  \vspace{-4mm}
  \caption{\fontsize{8}{9}\selectfont Sentiment flow for a discussion with \textbf{unresolved} dispute about the definition of ``white people" (top) and a discussion with \textbf{resolved} dispute on merging articles about van cat (bottom). The labels $\{\mathrm{NN, N, O, P, PP}\}$ are mapped to $\{-2, -1, 0, 1, 2\}$ in sequence. Sentiment values are convolved by using a Gaussian smoothing kernel, and then cubic-spline interpolation is conducted. Different speakers are represented by curves of different colors. Dashed vertical lines delimit turns. Representative sentences are labeled with letters and their sentiment labels are shown on the right. For unresolved dispute (top), we see that negative sentiment exists throughout the discussion. Whereas, for the resolved dispute (bottom), less negative sentiment is observed at the end of the discussion; participants also show appreciation after the problem is solved (e.g.\ E and F in the plot).}
  \vspace{-3mm}
  \label{fig:sentiment_flow}
\end{figure*}

\subsection{Dispute Detection}
We model dispute detection as a standard binary classification task, and investigate four major types of features as described below.

\noindent
\textbf{Lexical Features.} 
We first collect \texttt{unigram} and \texttt{bigram} features for each discussion.

\noindent
\textbf{Topic Features.} 
Articles on specific topics, such as politics or religions, tend to arouse more disputes. We thus extract the \texttt{category} information of the corresponding article for each talk page. We further utilize \texttt{unigrams} and \texttt{bigrams} of the category as topic features.

\noindent
\textbf{Discussion Features.} 
This type of feature aims to capture the structure of the discussion. Intuitively, the more turns or the more participants a discussion has, the more likely there is a dispute. Meanwhile, participants tend to produce longer utterances when they make arguments. We choose \texttt{number of turns}, \texttt{number of participants}, \texttt{average number of words in each turn} as features. In addition, the frequency of revisions made during the discussion has been shown to be good indicator for controversial articles~\cite{Vuong:2008:RCW:1341531.1341556}, that are presumably prone to have disputes. Therefore, we encode the \texttt{number of revisions} that happened during the discussion as a feature.

\noindent
\textbf{Sentiment Features.} 
This set of features encode the sentiment distribution and transition in the discussion. We train our sentiment tagging model on the full AAWD dataset, and run it on the Wikipedia dispute corpus. 

Given that consistent negative sentiment flow usually indicates an ongoing dispute, we first extract features from \texttt{sentiment distribution} in the form of \texttt{number/probability of sentiment per type}. 
We also estimate the \texttt{sentiment transition} probability $P(S_{t}\rightarrow S_{t+1})$ from our predictions, where $S_{t}$ and $S_{t+1}$ are sentiment labels for the current sentence and the next. 
We then have features as \texttt{number/portion of sentiment transitions per type}.

Features described above mostly depict the \textit{global} sentiment flow in the discussions. We further construct a \textit{local} version of them, since sentiment distribution may change as discussion proceeds. For example, less positive sentiment can be observed as dispute being escalated. 
We thus split each discussion into three equal length stages, and create sentiment distribution and transition features for each stage. 


\begin{table}[th]
\centering
    {\fontsize{8}{9}\selectfont
	\begin{tabular}{|l|cccc|}
    \hline
	& \textbf{Prec} & \textbf{Rec} & \textbf{F1} & \textbf{Acc}\\ \hline
	Baseline (Random) & 50.00 & 50.00 & 50.00& 50.00\\ 
	Baseline (All dispute) & 50.00 & 100.00 & 66.67 & 50.00\\ \hline
	Logistic Regression & 74.76 & 72.29 & 73.50 & 73.94\\ 
	SVM$_{Linear}$& 69.81 & 71.90 & 70.84 & 70.41\\ 
	SVM$_{RBF}$ & \textbf{77.38} & 79.14 & \textbf{78.25} & \textbf{80.00}\\ \hline
	\end{tabular}
	}
	\vspace{-4mm}
    \caption{\fontsize{8}{9}\selectfont Dispute detection results on Wikipedia Talk pages. The numbers are multiplied by 100. The items in \textbf{bold} are statistically significantly higher than others in the same column (paired-$t$ test, $p<0.05$). SVM with the RBF kernel achieves the best performance in precision, F1, and accuracy.}
    \label{tab:dispute_main}
\end{table}

\begin{table}[th]
\centering
    {\fontsize{8}{9}\selectfont
	\begin{tabular}{|l|cccc|}
    \hline
	& \textbf{Prec} & \textbf{Rec} & \textbf{F1} & \textbf{Acc}\\ \hline
 	Lexical (Lex) & 75.86 & 34.66 & 47.58 & 61.82\\ 
 	Topic (Top) & 68.44& 71.46 & 69.92 & 69.26 \\
	Discussion (Dis)	 & 69.73 & 76.14 & 72.79 & 71.54\\
  	Sentiment (Senti$_{g+l}$) & 72.54 & 69.52 & 71.00 & 71.60\\
  	
  	Top + Dis & 68.49 & 71.79 & 70.10 & 69.38\\
	Top + Dis + Senti$_{g}$ 	& 77.39 & 78.36 & 77.87 & 77.74\\
	Top + Dis + Senti$_{g+l}$ & 77.38 & \textit{79.14} & \textit{78.25} & \textbf{80.00}\\
	Lex + Top + Dis + Senti$_{g+l}$ & \textit{78.38} & 75.12 & 76.71 & 77.20\\ \hline
	\end{tabular}
	}
	\vspace{-4mm}
    \caption{\fontsize{8}{9}\selectfont Dispute detection results with different feature sets by SVM with RBF kernel. The numbers are multiplied by 100. Senti$_{g}$ represents global sentiment features, and Senti$_{g+l}$ includes both global and local features.
    The number in \textbf{bold} is statistically significantly higher than other numbers in the same column (paired-$t$ test, $p<0.05$), and the \textit{italic} entry has the highest absolute value.}
    \label{tab:dispute_feature}
\end{table}

\noindent
\textbf{Results and Error Analysis.} 
We experiment with logistic regression, SVM with linear and RBF kernels, which are effective methods in multiple text categorization tasks~\cite{Joachims:1999:MLS:299094.299104,Zhang:2001:TCB:593962.594015}. We normalize the features by standardization and conduct a 5-fold cross-validation. Two baselines are listed: (1) labels are randomly assigned; (2) all discussions have disputes. 

Main results for different classifiers are displayed in Table~\ref{tab:dispute_main}. All learning based methods outperform the two baselines, and among them, SVM with the RBF kernel achieves the best F1 score and accuracy (0.78 and 0.80). 
Experimental results with various combinations of features sets are displayed in Table~\ref{tab:dispute_feature}. As it can be seen, sentiment features obtains the best accuracy among the four types of features. A combination of topic, discussion, and sentiment features achieves the best performance on recall, F1, and accuracy. Specifically, the accuracy is significantly higher than all the other systems (paired-$t$ test, $p<0.05$).

After a closer look at the results, we find two main reasons for incorrect predictions. Firstly, errors from sentiment prediction get propagated into dispute detection. Due to the limitation of existing general-purpose lexicons, some opinionated dialog-specific terms are hard to catch. For example, ``I told you over and over again..." strongly suggests a negative sentiment, but no single word shows negative connotation. Constructing a lexicon tuned for conversational text might further improve the performance. Secondly, some dispute discussions are harder to detect than the others due to different dialog structures. For instance, the recalls for dispute discussions of ``controversy", ``RFC", and ``resolved" are 0.78, 0.79, and 0.86 respectively. We intend to design models that are able to capture dialog structures, such as pragmatic information, in the future work.

\noindent
\textbf{Sentiment Flow Visualization.} 
We visualize the sentiment flow of two disputed discussions in Figure~\ref{fig:sentiment_flow}.  
The plots reveal persistent negative sentiment in unresolved disputes (top). For the resolved dispute (bottom), participants show gratitude when the problem is settled.


%% file: conclusion.tex
We present a sentiment analysis-based approach to online dispute detection. We create a large-scale dispute corpus from Wikipedia Talk pages to study the problem. A sentiment prediction model based on isotonic CRFs is proposed to output sentiment labels at the sentence-level. Experiments on our dispute corpus also demonstrate that classifiers trained with sentiment tagging features outperform others that do not.